%% file: main.tex
\documentclass{article} 
\usepackage[preprint]{acl}

\input{math_commands.tex}

\usepackage{hyperref}
\usepackage{url}
\usepackage{xcolor}
\usepackage{algorithm}
\usepackage{algorithmic}
\usepackage{amsmath}
\usepackage{amssymb}
\usepackage{booktabs}
\usepackage{microtype}
\usepackage{inconsolata}
\usepackage{latexsym}
\usepackage[T1]{fontenc}
\usepackage[utf8]{inputenc}
\usepackage{pifont}
\usepackage{enumitem}
\usepackage{graphicx}
\usepackage{subcaption}
\usepackage{makecell}
\usepackage{tcolorbox}
\usepackage{caption}
\usepackage{listings}
\usepackage{textcomp}
\tcbuselibrary{breakable,listings}

\newcommand{\anna}[1]{{\color{black}#1}}
\newcommand{\nikita}[1]{{\color{black}#1}}
\newcommand{\seiji}[1]{{\color{black}#1}}

\newcommand{\method}[1]{BRIA}
\newcommand{\methodfull}[1]{BRanch–Inspect–Adopt}
\newcommand{\toolcount}[1]{1897}

\definecolor{deepgreen}{rgb}{0.0, 0.5, 0.0}
\definecolor{sigdrop}{rgb}{1.0, 0.5, 0.5}
\definecolor{moddrop}{rgb}{1.0, 0.8, 0.8}
\definecolor{slidrop}{rgb}{1.0, 0.9, 0.9}

\newcommand{\xmark}{\textcolor{red}{\textbf{\ding{55}}}}
\newcommand{\cmark}{\textcolor{deepgreen}{\textbf{\ding{51}}}}

\newtcolorbox[list inside=prompt,auto counter,number within=section]{prompt}[1][]{
    fontupper=\ttfamily\footnotesize,
    boxsep=5pt,
    left=0pt,
    right=0pt,
    top=0pt,
    bottom=0pt,
    boxrule=1pt,
    #1,
}

\title{Explore Separately, Adopt Selectively: Execution-Guided Tool Use over Large Tool Libraries}
\author{
  Anna Mazhar\thanks{Work done during internship at Megagon Labs.}\thanks{Equal contribution.} \\
  Cornell University \\
  \texttt{annam@cs.cornell.edu}
  \And
  Seiji Maekawa\footnotemark[2] \\
  Megagon Labs \\
  \texttt{seiji@megagon.ai}
  \And
  Nikita Bhutani \\
  Megagon Labs \\
  \texttt{nikita@megagon.ai}
}
\date{August 2026}

\begin{document}

\maketitle

\begin{abstract}
    Large language model agents must discover relevant tools from large libraries, e.g., thousands of tools, as new needs arise during task execution.
    When several candidates could satisfy the same need, execution may be required to identify the relevant one. Investigating candidates in a shared trajectory can leave alternatives unexplored or allow failed attempts to influence subsequent reasoning.
    To address this, we present \method{} (\methodfull{}), a simple yet effective agent that temporarily isolates candidate investigations in separate branches. Each branch retrieves and executes tools to investigate its candidate; a \textit{task agent} inspects the resulting execution evidence and adopts only useful results into its continuing trajectory. \method{} retrieves additional tools as new capability needs arise during task execution.
    We evaluate \method{} on PlanBench-XL and MCP-Atlas using a shared pool of \toolcount{} tools, three language models, and static and adaptive retrieval baselines. Across all settings, \method{} matches or exceeds the strongest retrieval baseline. On MCP-Atlas, it achieves an average claim score of 89.77, compared with 75.40 for MCP-Zero, a gain of 14.37 percentage points. \method{} also lies on the Pareto frontier for task performance and input-token cost.


\end{abstract}

\begin{figure}[t]
  \centering
  \includegraphics[width=\linewidth]{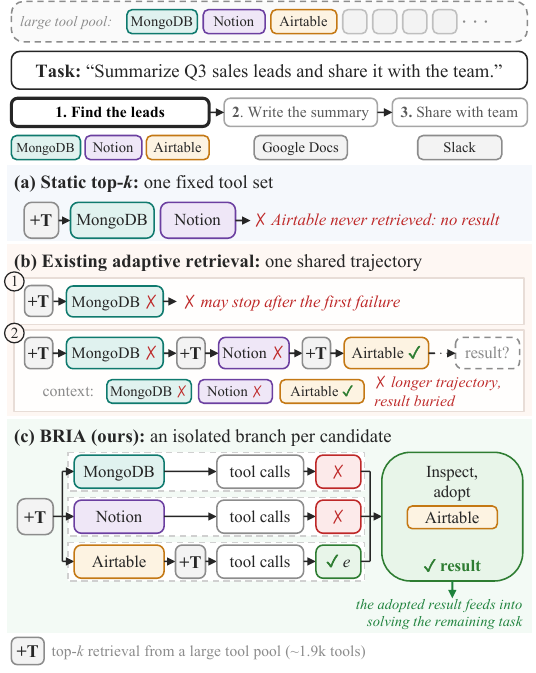}
  \caption{
    (a) Static top-$k$ retrieval may miss the right source. (b) Adaptive retrieval tries sources sequentially in one trajectory: it may stop after the first failure (\ding{172}) or accumulate failed attempts in context (\ding{173}). (c) \method{} explores each source in an isolated branch and adopts only the one grounded in execution evidence ($e$).}
  \label{fig:motivation}
\end{figure}

\begin{table*}[t]
  \centering
  \small
\scalebox{0.95}{
  \begin{tabular}{lcccc}
  \toprule
  \textbf{Method} &
  \textbf{Iterative Retrieval} &
  \textbf{Branching} &
  \textbf{Selective Adoption} &
  \textbf{Plug-and-Play} \\
  \midrule
    One-shot Top-$k$
      & \xmark & \xmark & \xmark & \cmark \\
    Tree-Planner \citep{treeplanner-iclr} & \xmark & \cmark & \xmark & \cmark \\
    MCP-Zero \citep{mcp-zero}
      & \cmark & \xmark &  \xmark & \cmark \\
    FitText \citep{zheng2026fittext}
      & \cmark & \cmark & \xmark & \cmark \\
    ToolGate \citep{toolgate}
      & \cmark & \xmark &  \cmark & \cmark \\
    ToolQP \citep{fang-glass-2026-beyond} & \cmark & \xmark &  \xmark & \xmark \\
    \midrule
    \textbf{BRIA (ours)}
      & \cmark & \cmark & \cmark & \cmark \\
    \bottomrule
  \end{tabular}
  }
  \caption{\nikita{Comparison of adaptive retrieval approaches. ``Branching'' denotes explicit investigation of candidate tool-use trajectories;
``Selective adoption'' denotes tool execution and selective retention of results from candidate investigations;
and ``Plug-and-Play'' indicates whether an arbitrary LLM can be used without parameter updates.}}
  \label{tab:method-comparison}
\end{table*}

\section{Introduction}
\label{sec:intro}

Large language model (LLM) agents increasingly rely on external tools such as APIs, databases, repositories and services to complete tasks that require capabilities beyond the model itself. As tool libraries grow to hundreds or thousands of tools \citep{metatool,toolomni,anytool,mcp-zero}, exposing the entire library at once becomes impractical. Retrieving too many tools increases context size and decision complexity, while retrieving too few can hide capabilities needed later in the task. Tool retrieval is therefore central to scalable tool use, much like retrieval-augmented generation, where retrieval must balance coverage and context size \citep{lewis2020retrieval,maekawa-etal-2024-retrieval,taguchi-etal-2025-efficient}.

\nikita{Recent work adapts tool retrieval as the needs of the agent evolve \cite{mcp-zero,sing2026,toolomni,hyperagent,zheng2026fittext}. However, adaptive retrieval can face \textit{tool uncertainty}: multiple candidate sources may offer the same capability, while only some contain the information relevant to the task. For example, a task may require information from an unspecified data source, with MongoDB, Notion databases, and Airtable all plausible candidates (Figure~\ref{fig:motivation}). The agent must investigate the candidates to identify the relevant source.}

\nikita{An adaptive retrieval method such as MCP-Zero \citep{mcp-zero}  may investigate candidates sequentially within a single trajectory. However, if an attempt is unsuccessful, the agent may stop before exploring other candidates. The failed attempt may also remain in the agent's context, increasing its size and potentially influencing subsequent reasoning. These risks are especially relevant to tasks involving long tool-call sequences, where context can grow substantially \citep{maekawa2026towards}.}

\nikita{
Existing methods address different aspects of this challenge. Tree-Planner and FitText explore alternative tool-use trajectories, while ToolGate supports  verified tool execution and state updates \citep{treeplanner-iclr,sing2026,toolgate,zheng2026fittext}. However, as Table~\ref{tab:method-comparison} shows, no existing method combines iterative retrieval, isolated candidate investigations, and selective adoption of execution results.}

\nikita{We introduce \method{} (\methodfull{}), an execution-grounded approach that brings these capabilities together for tool use over large libraries. When tool uncertainty arises because multiple candidates could satisfy a capability need, \method{} investigates each in an isolated branch, where tools can be retrieved and executed independently as needs evolve. The task agent inspects the resulting evidence, adopts useful results, and resumes the task. This process repeats as new capability needs arise.}


We evaluate \method{} on PlanBench-XL \citep{liu2026planbenchxlevaluatinglonghorizonplanning} and MCP-Atlas \citep{bandi2026mcpatlaslargescalebenchmarktooluse}, using a shared retrieval layer with 1897 tools, three LLM families, and four baseline methods. Across both benchmarks, \method{} matches or exceeds the strongest adaptive retrieval baseline. On PlanBench-XL, it achieves equal or higher success rates than MCP-Zero \cite{mcp-zero} while using less than half as many input tokens.  On MCP-Atlas, it improves task performance by up to 14.4 percentage points over the strongest baseline.  Our ablation studies and analyses further quantify the effects of isolated investigation and selective adoption, and show that branching is most useful when only a subset of candidates is needed. Our case studies illustrate how \method{} investigates candidates as tool uncertainty arises.


Our contributions are summarized as follows:
\begin{itemize}
    \item \nikita{We characterize \textit{tool uncertainty} in adaptive retrieval, where multiple candidates may satisfy the same capability need, and execution may be needed to identify which contains relevant information.}
    \item \nikita{We introduce \method{}, which isolates candidate investigations in separate branches and selectively adopts useful, execution-grounded results (Section~\ref{sec:proposal}).}
    \item  \nikita{We evaluate \method{} using two tool-use benchmarks and three LLMs against four baselines, and analyze its performance and input-token cost (Section~\ref{sec:experiments}).}
    \item \nikita{We analyze when branching helps and provide case studies of candidate investigation through tool execution (Section~\ref{ssec:case_study}).}
\end{itemize}

\begin{figure*}[t]
  \centering
  \includegraphics[width=\textwidth]{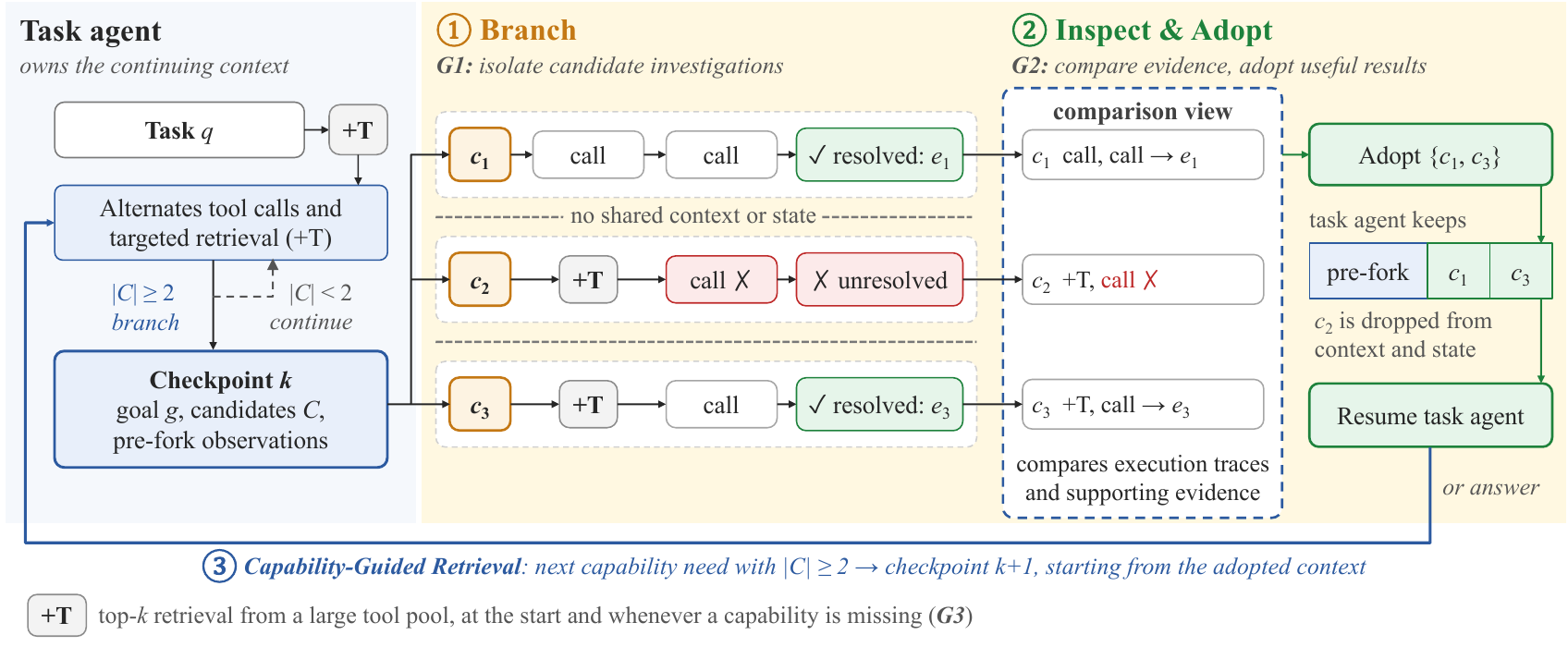}
  \caption{The \method{} workflow. The task agent opens a checkpoint when $|C| \geq 2$
candidates need the same investigation. \ding{172}~Each candidate is investigated
in an isolated branch (G1). \ding{173}~The task agent compares branch evidence and
adopts only useful branches (G2). \ding{174}~It resumes from the adopted context,
retrieving tools (+T) whenever a capability is missing (G3), and opens the next
checkpoint when another capability need has $|C| \geq 2$ candidates.}

  \label{fig:overview}
\end{figure*}

\section{Preliminaries}
\label{sec:prelim}

\nikita{
Let $q$ be a task and $\mathcal{T}_{\mathit{full}}$ a large tool library. An LLM agent, referred to as the \textit{task agent}, solves $q$ through iterative tool use but cannot inspect the full library directly.
Given a retrieval query $\rho$, a retriever $\mathcal{R}_k(\rho)$ ranks tools using their public schemas and returns at most $k$ tools for the task agent. The agent can retrieve tools again as new capability needs arise.

We study \textit{tool uncertainty}: a capability need may have multiple candidate tools whose schemas appear suitable, while tool execution is needed to determine which candidates provide relevant information. 
Our approach is guided by three design goals:
\begin{itemize}
    \item \textbf{G1: Isolate candidate investigations.}  Investigating one candidate should not affect the reasoning or execution state of another.
    \item \textbf{G2: Selectively adopt execution-grounded results.} 
         The task agent should inspect candidate-specific execution evidence and retain only results useful for continuing the task.
    \item \textbf{G3: Retrieve tools as capability needs arise.} 
     The task agent should retrieve additional tools when its currently available tools cannot satisfy a capability need.
\end{itemize}


}

\section{\method{}: \methodfull{}}
\label{sec:proposal}
\seiji{
Figure~\ref{fig:overview} illustrates how \method{} implements these design goals. Its core loop combines branching, inspection/adoption, and capability-guided retrieval. 
This allows the agent to efficiently explore multiple candidate solutions while ensuring that each investigation remains isolated and evidence is properly grounded.
}


\subsection{Overview of BRIA}
\label{ssec:parent}

\nikita{
\method{} combines an LLM-based \textit{task agent} with a controller.  The task agent determines how to advance the task and which candidate results to retain. The controller manages tool retrieval and execution, creates isolated branches, and presents their outcomes to the task agent for inspection.

The task agent begins with the task and an initial set of retrieved tools, then alternates between tool calls and targeted retrieval as new capability needs arise. When at least two concrete candidates require independent investigation of the same capability need, the task agent requests a \textit{checkpoint}. The checkpoint records the investigation goal $g$, candidate set $C$, and successful pre-fork observations. 
The controller creates an isolated branch for each candidate. Each branch can retrieve and execute tools to investigate its candidate without seeing the actions or results of sibling branches. After the investigations finish, it presents their outcomes to the task agent, which inspects the evidence and selects results to retain. Unadopted investigations remain in the audit log and do not influence subsequent reasoning. The task agent then resumes its trajectory. This process can repeat when another capability need has multiple candidates.}




\subsection{Branch (G1)}
\label{ssec:branching}

The task agent opens a checkpoint when at least two concrete candidates require independent investigation of the same capability need. It investigates these candidates as alternative sources, not as alternative plans for completing the overall task.

For each candidate $c_i \in C$, the controller creates an isolated branch $b_i$.  Each branch receives the original task, the shared investigation goal, its assigned candidate, the callable tools available at the checkpoint, and a fixed record of pre-fork tool executions. The record contains execution arguments, parsed results, and failure details, but not the task agent's reasoning. After the fork, a branch can retrieve additional tools independently. It sees its own actions and results, but cannot access actions or observations from sibling branches.

A branch is \textit{resolved} if it returns a conclusion and cites at least one successful post-fork observation produced by that branch. Otherwise, it remains \textit{unresolved}, for example when evidence is insufficient or failures prevent the investigation from proceeding. The controller verifies the cited observation's provenance, but does not determine whether it answers the investigation goal.




\subsection{Inspect \& Adopt (G2)}
\label{ssec:inspection}

\subsubsection{Inspect: Evidence Comparison}
\nikita{After all branch investigations finish, the controller presents their outcomes to the task agent in a temporary comparison view. For each candidate, the view includes its resolution status, execution trace (tool calls, arguments, results, and failures), conclusion, and cited evidence. The task agent checks the traces and evidence rather than relying on results alone, since similar results can come from different or incomplete executions. The temporary view exposes all branch outcomes without adding their full histories to the task agent's continuing context.}


\subsubsection{Adopt: Selective Integration}
After inspecting the comparison view, the task agent selects which branch investigations to retain. It may adopt one or more branches, or none, depending on the evidence and the task's needs. For each adopted branch, the controller makes its successful observations, retrieved tools, and execution history available to the task agent. 

Adoption determines which investigation records remain available in the continuing trajectory: what the task agent can see, which observations it can cite, and which prior executions can support later tool calls. After adoption, the controller reconstructs the continuing context and execution history from the pre-fork history and adopted investigations. Unadopted investigations remain in the audit log and are excluded from subsequent reasoning and tool calls. Adoption does not certify semantic correctness. The task agent then resumes its ordinary tool use and may open another checkpoint as new capability needs arise.

\subsection{Capability-Guided Retrieval (G3)}
\label{ssec:retrieval_recovery}



\seiji{
Retrieval provides tools for advancing the task and investigating candidates. The initial retrieval uses the original task text. As the task progresses, the task agent identifies a missing capability and represents it as $\nu=(\textit{goal},\textit{available inputs},\textit{required output},\textit{domain})$. The controller converts this need into a query for the shared dense index. 
If the retrieved tools do not satisfy the need, the agent may retry the retrieval with an updated query or repeat the exact same query to explore additional tools.
In cases of repeated retrieval, the controller excludes tools already available to the agent and returns the next-best unseen tools.

If a tool call fails because a required input is unavailable, the controller returns the failure as recovery context. The task agent inspects the failed call, the tool schema, and prior observations to identify the missing dependency. It then uses an available tool or retrieves the capability needed to obtain that input before retrying the blocked step. 
After a checkpoint, each branch maintains its own retrieval history, allowing retrieval to adapt independently to that branch's needs.
}



\subsection{Execution Isolation and Guarantees}
\label{ssec:lineage}

Context isolation alone is insufficient in a stateful environment. The controller associates each tool call with its execution lineage and reconstructs the state available to the task agent or a branch from that lineage. Each branch inherits the task agent's pre-fork state but cannot access state produced by sibling branches. After adoption, the task agent's state consists of its pre-fork history and the execution history of adopted branches. Unadopted work cannot affect subsequent execution or receive task credit. The controller also enforces schema-valid tool calls, evidence ownership, shared execution budgets, and the use of non-mutating tools within branches. 

\section{Experiments}
\label{sec:experiments}



{
We organize the evaluation around four research questions.
\textbf{(RQ1: Effectiveness)} Can \method{} improve task performance over baseline methods in the shared tool pool?
\textbf{(RQ2: Cost)} Can \method{} achieve higher task performance at lower token cost?
\textbf{(RQ3: Ablation)} How does each component of \method{} affect performance and cost?
\textbf{(RQ4: Analysis)} When and why does \method{} help?
We additionally present case studies to illustrate how \method{} operates in practice.
}

\subsection{Setup}
\paragraph{Benchmarks}
We evaluate on PlanBench-XL \citep{liu2026planbenchxlevaluatinglonghorizonplanning} and MCP-Atlas \citep{bandi2026mcpatlaslargescalebenchmarktooluse} since they provide complementary tool use settings.
{PlanBench-XL} is a synthetic benchmark with 327 retail-domain tasks requiring long-horizon tool composition in noisy, large-scale environments with blocker tools.
{MCP-Atlas} contains 1,000 tasks across 36 real MCP servers. We use the public subset and read-only tools from MCP-Atlas, yielding 500 tasks and 232 tools. Unlike PlanBench-XL, MCP-Atlas uses realistic tool schemas, server interactions and execution behavior. We provide further details in Appendix~\ref{app:benchmark_details}.


\paragraph{Retrieval Setup}
\label{sec:common_retrieval}
We place both benchmarks behind a shared retrieval layer to evaluate them under a common tool-discovery setting. We combine 1665 PlanBench-XL tools and 232 MCP-Atlas tools into a pool of \toolcount{} tools. 
We use {Qwen3-Embedding-8B} \citep{zhang2025qwen3embedding} for embedding retrieval queries and tool documents and rank tools by cosine similarity. The retrieval and route-planning components are shared across benchmarks, while execution and evaluation adapters remain benchmark-specific. All methods use the same bi-encoder retrieval setup.




\begin{table*}[!t]
    \centering
    \small
    \scalebox{1}{
    \begin{tabular}{@{}lrrrrrr@{}}
    \toprule
    & \multicolumn{3}{c}{PlanBench-XL} & \multicolumn{3}{c}{MCP-Atlas} \\
    \cmidrule(lr){2-4}\cmidrule(lr){5-7}
    Method & \shortstack{GPT-6\\Luna} & \shortstack{Gemini-3.8\\Flash} & \shortstack{GLM-5.3\\Flash} & \shortstack{GPT-6\\Luna} & \shortstack{Gemini-3.8\\Flash} & \shortstack{GLM-5.3\\Flash} \\
    \midrule
    \shortstack[l]{Task-Gold-Only} & 76.00 & 100.00 & 100.00 & 66.48 & 72.15 & 78.14 \\
    \midrule
    One-shot $k=50$ & 0.00 & 0.00 & 0.00 & 48.27 & 56.64 & 64.10 \\
    One-shot $k=250$ & 8.00 & 10.67 & 10.67 & 54.51 & 59.93 & 66.71 \\
    One-shot $k=500$ & 18.67 & 30.67 & 22.67 & 60.39 & 59.07 & 71.50 \\
    MCP-Zero & 18.67 & \textbf{37.33} & 32.00 & 60.47 & 75.40 & 78.67 \\
    FitText & 2.67 & 10.67 & 12.00 & 24.68 & 70.56 & 64.92 \\
    ToolGate & 0.00 & 0.00 & 0.00 & 24.43 & 37.32 & 39.67 \\
    \midrule
    \textbf{BRIA} & \textbf{22.67} & \textbf{37.33} & \textbf{38.67} & \textbf{62.35} & \textbf{89.77} & \textbf{81.54} \\
    \bottomrule
    \end{tabular}
    }
    \caption{Overall scores. PlanBench-XL: answer-correctness $\wedge$ solution-path-coverage; MCP-Atlas: average claim score. Bold numbers indicate the best performance.}
    \label{tab:dev75-results}
\end{table*}

\paragraph{Baselines}
\label{sec:eval_baselines}
We compare \method{} with adaptive tool-discovery approaches \textbf{MCP-Zero} \citep{mcp-zero}, \textbf{FitText} \citep{zheng2026fittext}, and \textbf{ToolGate} \citep{toolgate}, which revise or expand the available tool set during execution. We use the official implementations for MCP-Zero\footnote{\url{https://github.com/xfey/MCP-Zero}} and ToolGate\footnote{\url{https://github.com/OceannTwT/ToolGate}} and re-implement FitText based on the paper.
We additionally compare against a conventional static-retrieval baseline,  \textbf{One-shot top-$k$}, which retrieves the top-$k$ tools once at task start and exposes them to a standard ReAct agent \citep{yao2023react}. We evaluate $k\in\{50,250,500\}$. Finally, we include \textbf{Task-Gold-only} as a reference setting in which a standard LLM agent is equipped only with task-relevant gold tools and no blocker tools. 


\paragraph{Models}
We employ three LLMs from different model families including proprietary and open-source models: GPT-6-Luna \citep{openai2026gpt6}, Gemini-3.8-Flash \citep{gemini3.8-2026}, and GLM-5.3-Flash \citep{glm5team2026glm5vibecodingagentic}.
We set temperature to 0 when models allow and use the medium reasoning effort for all models. 
The prompts used for the experiments are shown in Appendix \ref{app:ssec:prompts}.

\paragraph{Evaluation}
\label{sec:eval_metrics}
We sample 75 representative tasks from each benchmark due to the breadth of experiments across three models and seven methods. 
We use a retrieval unit of $k=50$ for all methods; other configuration details are in Appendix~\ref{app:config}. Our primary metric is task success. For PlanBench-XL, a task succeeds when the final response contains the answer and all gold tools are correctly invoked. For MCP-Atlas, we report the official claim-coverage score from its claim-level evaluator, using GPT-6-Luna \citep{openai2026gpt6} as the LLM judge.



\subsection{Main Results (RQ1)}
\label{ssec:main_results}

\begin{figure}[t]
    \centering
    \begin{subfigure}[t]{0.48\linewidth}
        \centering
        \includegraphics[width=\linewidth]{
            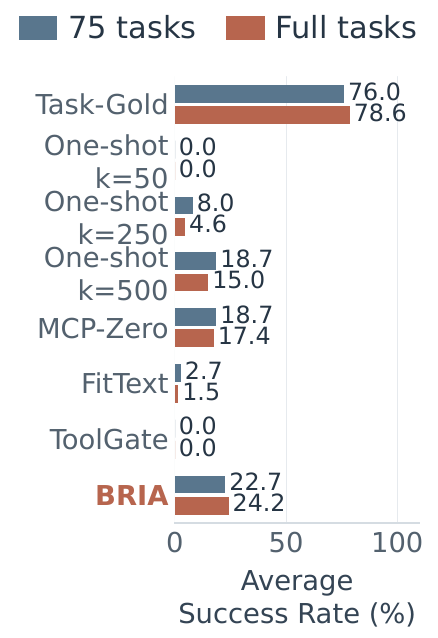}
        \caption{PlanBench XL}
        \label{fig:planbench-dev75-vs-full}
    \end{subfigure}
    \hfill
    \begin{subfigure}[t]{0.48\linewidth}
        \centering
        \includegraphics[width=\linewidth]{
            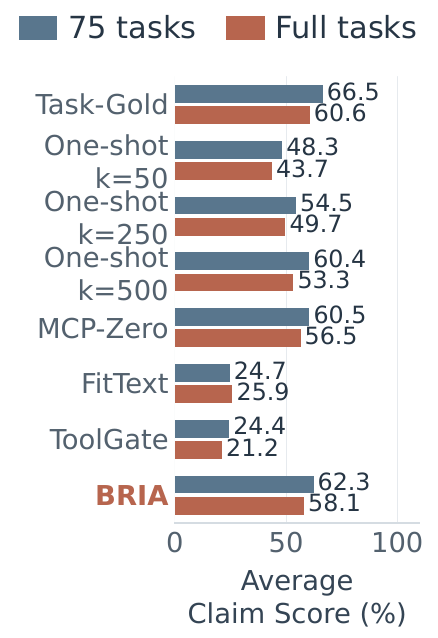}
        \caption{MCP-Atlas}
        \label{fig:mcp-atlas-dev75-vs-full}
    \end{subfigure}
    \caption{Comparison of sampled and full-set scores with GPT-6-Luna.}
    \label{fig:dev75-vs-full}
\end{figure}

{
Table \ref{tab:dev75-results} compares \method{} with the baselines on PlanBench-XL and MCP-Atlas.

\textbf{BRIA consistently achieves the best score across benchmarks and models. }
We observe 6.67\% and 14.37\% score improvements over the second best method on PlanBench-XL and MCP-Atlas, respectively. The gains are particularly notable over one-shot retrieval, which lacks the adaptive retrieval and route exploration capabilities used by \method{} to handle blocker and noisy tools. Increasing $k$ improves performance, but it remains substantially below \method{} while exposing more tools and incurring additional computational costs.


\textbf{FitText suffers from over-searching, ToolGate struggles with type-compatible tools.}
FitText frequently exhausts its search budget before arriving at a solution, doing so in 52--68 of the 75 PlanBench-XL runs. ToolGate struggles when multiple tools have compatible input and output types, as it selects tools based primarily on type compatibility without considering the broader task context.
{It exhausts its search budget in 33--68 runs, highlighting the challenge of selecting useful tools among plausible alternatives.}
{MCP-Zero is the strongest baseline: it is second-best in almost every setting. However, it follows a single trajectory at a time, so it cannot compare alternatives when a blocker tool appears plausible. \method{}'s branching addresses exactly this case.}

\textbf{\method{} can outperform Task-Gold-only in certain scenarios.}
On MCP-Atlas, \method{} outperforms Task-Gold-only with Gemini-3.8-Flash and GLM-5.3-Flash, even though the agent is provided with gold tools for each task.
This suggests that \method{}'s gains extend beyond identifying the right tools to using them effectively. Its explicit exploration and selective adoption of tool-use paths may also help avoid common execution errors, such as malformed actions and inefficient sequential tool use.
We observe a similar trend in PlanBench-XL when a small number of blocker tools are added to task-specific gold tools (see Section \ref{sssec:no-retrieval}). 


To verify that the results from our sampled tasks generalize to the full benchmarks,
we repeat the experiments on 325 tasks in PlanBench-XL and 500 tasks in MCP-Atlas using GPT-6-Luna.
Figure \ref{fig:dev75-vs-full} shows the results. 
\textbf{BRIA maintains its superior performance across the full benchmarks.} We additionally report results with bootstrapping confidence intervals in Appendix \ref{app:additional_experiments}.
}





\begin{figure}[t]
    \centering
    \begin{subfigure}[t]{\linewidth}
        \centering
        \includegraphics[width=\linewidth]{
            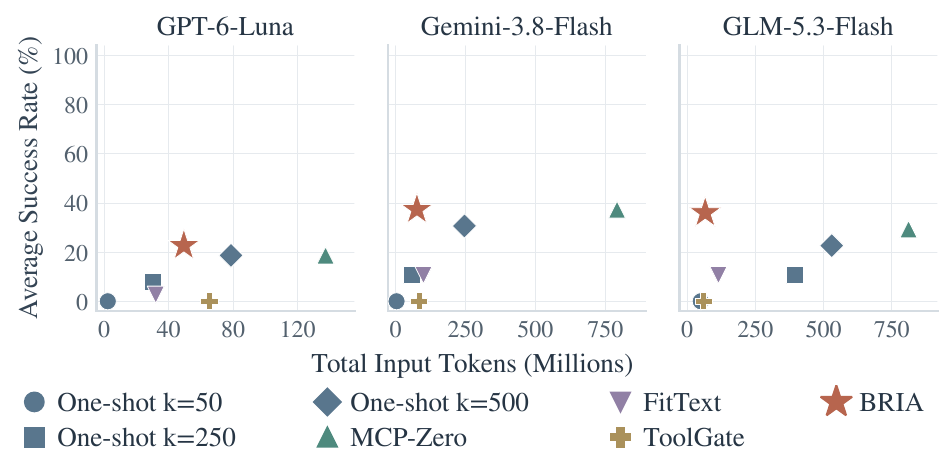}
        \caption{PlanBench XL}
        \label{fig:planbench-cost-vs-score}
    \end{subfigure}
    \hfill
    \begin{subfigure}[t]{\linewidth}
        \centering
        \includegraphics[width=\linewidth]{
            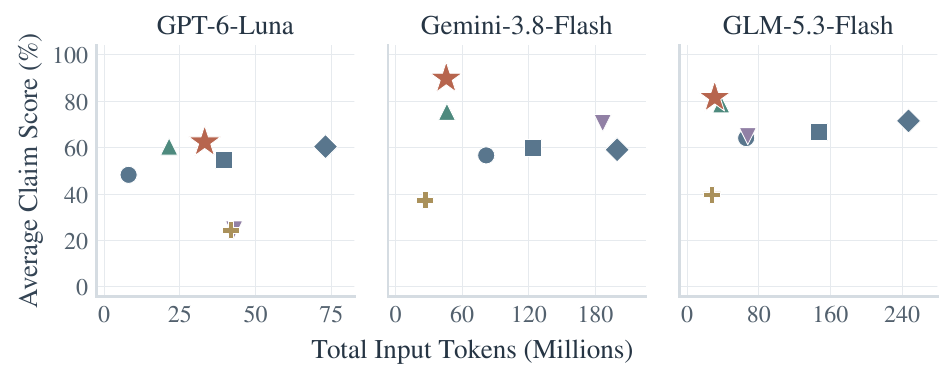}
        \caption{MCP-Atlas}
        \label{fig:mcp-atlas-cost-vs-score}
    \end{subfigure}
    \caption{Score vs total input token amount. }
    \label{fig:cost}
\end{figure}

\subsection{Cost (RQ2)}
\label{ssec:input_cost}

{
\textbf{\method{} lies on the accuracy--cost frontier in all six settings.} 
Figure \ref{fig:cost} shows the task score versus total input tokens. Across both benchmarks, \method{} achieves higher accuracy than One-shot $k{=}500$ while using fewer input tokens, demonstrating the benefit of adaptive tool retrieval over retrieving a large static tool set.
On PlanBench-XL, \method{} is more accurate than MCP-Zero while using less than half the input tokens. This demonstrates that it can effectively balance accuracy and costs in scenarios that require exploration of multiple plausible tools. 
On MCP-Atlas, \method{} likewise achieves a favorable accuracy--cost trade-off. We observe similar trends when analyzing using total input and output tokens (see Appendix \ref{app:additional_experiments}).

}



\subsection{Ablation Study (RQ3)}
\label{ssec:ablation}
\seiji{
We conduct an ablation study to understand the contributions of different components of BRIA to its overall performance and cost efficiency.

\begin{table}[t]
    \centering
    \small
    \setlength{\tabcolsep}{4pt}
    \begin{tabular}{@{}lrrr@{}}
    \toprule
    Method & \shortstack{GPT-6\\Luna} & \shortstack{Gemini-3.8\\Flash} & \shortstack{GLM-5.3\\Flash} \\
    \midrule
    Task-Gold-only & 76.00 & 100.00 & 100.00 \\
    \midrule
    Gold + Blockers & 48.00 & 96.00 & 92.00 \\
    BRIA w/o Retrieval & \textbf{88.00} & \textbf{98.67} & \textbf{97.33} \\
    \bottomrule
    \end{tabular}
    \caption{No-retrieval setup comparison on PlanBench-XL. Gold-only and Gold + Blockers use a standard LLM agent; BRIA w/o Retrieval uses the Gold + Blockers tool menu.}
    \label{tab:planbench-no-retrieval}
\end{table}

\subsubsection{No Retrieval Setup}
\label{sssec:no-retrieval}
To examine the robustness of BRIA's branching mechanism under perfect retrieval recall in the presence of blocker tools, we introduce an additional setting, Gold + Blockers, and test BRIA without the retrieval component. 
We use PlanBench-XL for this evaluation since it contains task-specific blockers that allow us to create this setting effectively. 
In the setting, each task includes the union of gold tools across solution paths and three blocker variants per gold tools. Tool menus contain an average of 53.39 tools, i.e., 13.35 gold tools and 40.04 blockers. 

\paragraph{BRIA w/o Retrieval outperforms standard LLM agents in the presence of blockers.}
Table \ref{tab:planbench-no-retrieval} shows the performance comparison of BRIA without retrieval against standard LLM agents under the no-retrieval setup. 
This demonstrates that BRIA's branching mechanism successfully resolves the tool selection uncertainty introduced by blocker tools.
Interestingly, with GPT-6-Luna, BRIA w/o retrieval outperforms Task-Gold-only with a standard LLM agent (88\% vs. 76\%).  \anna{Trace analysis shows that the standard agent often terminates prematurely (15 out of 18 failed tasks). It concludes that no available tool can complete the chain, even though all required tools are provided, whereas BRIA's branching surfaces such chains.
}

\subsubsection{BRIA without Inspection}
\label{sssec:no-inspection}
To assess the effectiveness of BRIA's inspection component, we also test BRIA without inspection, i.e., we automatically insert all the branches into the task agent context without any inspection. 

\paragraph{BRIA's inspection component contributes to both accuracy and cost efficiency.}
Table \ref{tab:no-inspection-accuracy-cost} shows the impact of removing the inspection component on both the success rate and the input token cost for PlanBench-XL and MCP-Atlas.
Since the inspection component keeps only the branches that are relevant to the task, it reduces the number of branches inserted into the task agent context, leading to lower input token costs in all cases. 
On MCP-Atlas, BRIA achieves better scores with fewer input tokens. 
Despite eliminating the explicit selection call, automatic all adoption does not yield lower costs. This is because inserting all branches increases the downstream processing cost due to retaining branch information.

\subsubsection{No Repeat-query for Unseen Tools}
\label{sssec:no-repeat-query}
To evaluate the impact of repeat-query for unseen tools, we test BRIA without this feature, i.e., the same query receives the same retrieval result. 

\paragraph{Repeat-query for unseen tools improves retrieval robustness.}
Table \ref{tab:no-repeat-query-novelty} shows the impact of disabling repeat-query for unseen tools on both the success rate of GPT-6-Luna for PlanBench-XL and MCP-Atlas.
While repeat-query for unseen tools is helpful for improving the performance, its impact is particularly pronounced in scenarios where the initial retrieval may miss relevant tools. 
This highlights the importance of allowing multiple retrieval attempts to ensure that the agent has access to all necessary tools for successful task completion.
}

\begin{table}[t]
  \centering
  \small
  \scalebox{0.88}{
  \begin{tabular}{@{}llrr@{}}
    \toprule
    Model & Method & Score (\%) & Input (M) \\
    \midrule
    \multicolumn{4}{c}{\textbf{PlanBench-XL (Success Rate)}} \\
    \addlinespace[2pt]
    GPT-6-Luna & BRIA & 22.67 & \textbf{49.45} \\
     & No Inspection & \textbf{24.00} & 52.20 \\
    \addlinespace[2pt]
    Gemini-3.8-Flash & BRIA & \textbf{37.33} & \textbf{76.23} \\
     & No Inspection & \textbf{37.33} & 79.29 \\
    \addlinespace[2pt]
    GLM-5.3-Flash & BRIA & \textbf{38.67} & \textbf{66.43} \\
     & No Inspection & \textbf{38.67} & 68.53 \\
    \midrule
    \multicolumn{4}{c}{\textbf{MCP-Atlas (Claim Score)}} \\
    \addlinespace[2pt]
    GPT-6-Luna & BRIA & \textbf{62.35} & \textbf{33.19} \\
     & No Inspection & 59.13 & 50.27 \\
    \addlinespace[2pt]
    Gemini-3.8-Flash & BRIA & \textbf{89.77} & \textbf{45.61} \\
     & No Inspection & 87.52 & 51.84 \\
    \addlinespace[2pt]
    GLM-5.3-Flash & BRIA & \textbf{81.54} & \textbf{30.82} \\
     & No Inspection & 74.60 & 33.15 \\
    \bottomrule
  \end{tabular}
  }
  \caption{BRIA vs. No Inspection on 75 tasks per benchmark. ``Input'' denotes the total input token amount. }
  \label{tab:no-inspection-accuracy-cost}
\end{table}

\begin{table}[t]
  \centering
  \small
  \scalebox{0.85}{
  \begin{tabular}{@{}lrr@{}}
    \toprule
    Method & \shortstack{PlanBench-XL\\Success (\%)} & \shortstack{MCP-Atlas\\Claim Score (\%)} \\
    \midrule
    BRIA & \textbf{22.67} & \textbf{62.35} \\
    \shortstack[l]{No Repeat-Query Novelty} & 16.00 & 60.51 \\
    \bottomrule
  \end{tabular}
  }
  \caption{Repeated-query novelty ablation with GPT-6-Luna on 75 tasks per benchmark.}
  \label{tab:no-repeat-query-novelty}
\end{table}

\subsection{Analysis (RQ4)}
\label{ssec:analysis}

\seiji{
We analyze the effectiveness of each component of \method{} on the full benchmarks with GPT-6-Luna.

\subsubsection{Branch Effectiveness}
\paragraph{Branching helps when only a subset of candidates is needed.}
BRIA opens a checkpoint on 24.5\% of PlanBench-XL tasks and 31.0\% of MCP-Atlas tasks.
On PlanBench-XL, it branches mostly on tasks in which one input leads to several alternative routes (50.6\%, compared with at most 18.2\% for other task types).
When such a checkpoint completes, \method{} succeeds on 36.0\% of these tasks, well above its overall 24.2\%.
On MCP-Atlas, where the tasks on which \method{} branches are harder than average, branching helps most when the task agent must find which of several databases, repositories, or records matches the request.
The task agent then adopts only the matching branches, typically about 40\% of them, and on these tasks \method{} wins 29 and loses 13 against MCP-Zero.
Across all runs, 87\% of branches resolve with a conclusion grounded in their own evidence, so the task agent decides on executed results rather than guesses.
Section~\ref{ssec:case_study} walks through one such task.

We observe that branching provides limited benefit for tasks that require results from all candidates. 
For example, when collecting coordinates for several places or prices for several dates, the task agent needs to adopt every branch. In such cases, branching adds overhead without the benefit of selective adoption.

\subsubsection{Capability-Guided Retrieval}

\paragraph{Capability-guided retrieval improves tool coverage and route visibility.}
\method{} retrieves new tools whenever the visible ones cannot advance the task, and whether it finds the right ones largely determines success: on PlanBench-XL, every method succeeds on at most 12\% of the tasks in which no complete valid route ever becomes visible.
One-shot $k{=}50$ never surfaces such a route, which explains its zero success.
\method{}'s capability-guided queries surface 57.0\% of the gold tools, compared with 44.0\% for MCP-Zero, and make a complete route visible on 83 of the 327 tasks, compared with 68.
Section~\ref{ssec:case_study} walks through one such task.
Once a complete route is visible, \method{} turns it into a correct answer more often than One-shot $k{=}500$ (Success rates: 81.9\% and 53.3\%, respectively). 
This demonstrates the benefit of limiting exposure to irrelevant tools.

Taken together, the components are complementary: capability-guided retrieval decides which tasks can be solved and keeps each call small, and branching-then-inspection helps when the agent must choose among candidates. 
Because branching is opt-in, it adds cost only where it is used.
}

\subsection{Case Study}
\label{ssec:case_study}
{To illustrate how \method{} handles multi-step tasks, we present two examples using GPT-6-Luna.

\subsubsection{Branching}
The following MCP-Atlas task illustrates how branching helps:
\begin{tcolorbox}
We have data [\ldots] I need to retrieve the title for a video game, the price, and what kind of purchase method they used on the same date as donald trump was named president of usa for the first term.
\end{tcolorbox}
The task does not specify where the relevant records are stored.
MCP-Atlas contains two databases, and its store database has six collections, several of which could contain purchase information. 
\method{} opens a checkpoint over the two databases. One branch investigates the movie database and finds no relevant records.
The other investigates the store database and identifies \emph{Customers} as containing the requested game title, price, purchase method, and date. The agent adopts the store branch's results and continues with that evidence.

In contrast, MCP-Zero investigates candidates sequentially. It queries the \emph{Purchase History} collection and finds a record dated January 20, 2017: \$192.61, paid in cash, with category ``Accessories.'' Because the record contains no game title, MCP-Zero concludes, \emph{``I can't determine which game was bought from the available records''}. The example shows how the first plausible record can fail to answer the task, while investigating another candidate reveals the relevant collection.

}

{
\subsubsection{Capability-guided Retrieval}

The following PlanBench-XL task illustrates how capability-guided retrieval helps:

\begin{tcolorbox}
An after-sales teammate [\dots] the shipment ID is \texttt{shp\_14090}, and the customer name is \texttt{Isaac Brooks}. [\dots] What refund completion time is recorded for that case?
\end{tcolorbox}

Answering the task requires a chain of five tools: shipment to order $\rightarrow$ return request $\rightarrow$ internal review $\rightarrow$ finance refund $\rightarrow$ refund time. Later tools are less related to the original request, making them harder to retrieve from the task text alone.


After each step, \method{} represents the next missing capability as a structured need and retrieves a suitable tool. For example, it requests a tool that maps a return audit ID to a refund transaction ID. This process continues through the chain until \method{} reaches the final \emph{refund-time} tool.

}


\section{Related Work}
\label{sec:related}
{
\paragraph{Large-scale tool retrieval}
AnyTool~\citep{anytool} searches large tool libraries hierarchically, while ToolScope~\citep{liu-etal-2026-toolscope} reduces redundancy among similar tools.
Re-Invoke~\citep{chen-etal-2024-invoke}, ToolDreamer~\citep{sengupta-etal-2026-tooldreamer}, and Multi-Field Tool Retrieval~\citep{tang2026multifieldtoolretrieval} improve request-to-tool alignment through query rewriting, hypothetical tool descriptions, or structured representations.
For multi-tool requirements, ToolQP~\citep{fang-glass-2026-beyond} decomposes complex requests into subtask queries, COLT~\citep{qu-2024-towards} promotes retrieval of a complete and diverse tool set, and HYSET~\citep{hong2026toolsislandssetleveltool} scores sets of mutually useful tools rather than treating each tool independently.
    
\paragraph{Adaptive tool discovery}
MCP-Zero~\citep{mcp-zero} iteratively requests additional tools when its current capabilities are insufficient, while SING~\citep{sing2026} retrieves tools according to evolving intentions and execution state. FitText~\citep{zheng2026fittext} repeatedly refines descriptions of the capabilities currently needed, while ToolOmni~\citep{toolomni} and
HyperAgent~\citep{hyperagent} expand retrieval around unresolved requirements. 

\paragraph{Tool planning and execution verification}
TOOLVERIFIER~\citep{toolverifier} distinguishes closely related tools through self-verification, while ToolGate~\citep{toolgate} verifies tool conditions and outputs before allowing them to modify the agent's trusted state. Search-based planners such as ToolChain*~\citep{toolchain_iclr}, Tree-Planner~\citep{treeplanner-iclr}, and ToolTree~\citep{tooltree} explore alternative action trajectories using execution feedback.

Together, these lines of work improve tool retrieval, adaptive discovery, planning, and execution verification. The remaining challenge we address is how to investigate competing candidates without allowing unselected investigations to affect the task agent's continuing context and execution history. \method{} addresses this challenge by letting the task agent compare candidate-specific execution evidence and retain only the investigations needed to continue the task. This limits the context carried forward from unsuccessful or irrelevant candidates while preserving selected observations and tool calls for subsequent reasoning.


}

\section{Conclusion}
\label{sec:conclusion}
{We presented \method{}, an agent for tool use over large tool libraries that handles tool uncertainty by investigating each candidate in an isolated branch and adopting only the results supported by execution evidence. On PlanBench-XL and MCP-Atlas with \toolcount{} tools and three LLMs, \method{} matches or exceeds the strongest baseline in every setting, by up to 14.4 points, while achieving a favorable performance--cost trade-off. Our analysis shows that branching is most useful when only a subset of candidates is needed.}

\section*{Limitations}
\anna{While \method{} improves tool use performance over large tool libraries, it has several limitations.
First, our experiments employ a shared bi-encoder-based retrieval without an additional reranking stage. Reranking is orthogonal to \method{}'s methodology including execution-guided retrieval and candidate branching. Evaluating their combination is left for future work.
Second, \method{} restricts branches to non-mutating tools so that one candidate investigation cannot affect another, and our MCP-Atlas evaluation accordingly uses only read-only tools. We plan to support tools that modify external state, such as placing an order, through mechanisms such as sandboxing or rollback of unadopted branches.
Third, for a fair comparison with the baselines, the task agent and all of its branches share the same fixed retrieval budget. We intend to explore adaptive budget allocation across branches, such as prioritizing more promising candidates or terminating unpromising branches early, which could further improve efficiency when a checkpoint involves many candidates.
Lastly, to cover seven methods and three LLMs, most of our experiments use 75 sampled tasks per benchmark, with full-benchmark experiments conducted with GPT-6-Luna, which confirm the trends observed on the sampled tasks (Section~\ref{ssec:main_results}). Evaluating \method{} on additional domains and tool ecosystems is an interesting direction for future work.}

\section*{Ethical Considerations}
\anna{\method{} helps LLM agents find and use the right tools in large tool libraries, and all of our experiments run on the public PlanBench-XL and MCP-Atlas benchmarks without private user data or credentials. The method itself raises no direct ethical issues, but the agents built on it may. Tool-using agents act on external systems such as databases, repositories, and online services, where a wrong tool call can trigger unintended actions or reveal sensitive information. \method{}'s read-only branches limit this risk during candidate investigation, and we recommend that deployments also use access control and human oversight for tools with side effects, so that the benefits of \method{} can be realized safely.

We used AI assistants, including ChatGPT and Copilot, to help with coding and with polishing the writing of this paper. All generated content was carefully reviewed and revised by the authors to ensure its accuracy and consistency with our research goals.}

\bibliography{ref}

\appendix

\section{Configuration}
\label{app:config}
Table \ref{tab:config} shows the detailed configuration for our experiments. 

\begin{table}[h]
    \centering
    \small
    \scalebox{0.85}{
    \begin{tabular}{@{}ll@{}}
        \toprule
        Setting & Value \\
        \midrule
        Embedding model & Qwen3-Embedding-8B \\
        Retrieval top-$k$ & 50 \\
        Max retrievals / task & 20 \\
        Max tool calls / task & 100 \\
        Max model calls / task & 200 \\
        Max internal actions / task & 200 \\
        Max candidates / checkpoint & 16 \\
        Nested checkpoints & not allowed \\
        Response format & JSON object if applicable \\
        \bottomrule
    \end{tabular}
    }
    \caption{Final \method{} configuration. All budgets are shared across the
    task agent and all branches of a task.}
    \label{tab:config}
\end{table}

\begin{table*}[h]
  \centering
  \resizebox{0.9\linewidth}{!}{
  \begin{tabular}{lrrlc}
      \toprule
      \textbf{Model} & \textbf{Size} & \textbf{Context} & \textbf{Model name / snapshot} & \textbf{License} \\
      \midrule
      GPT-6-Luna & --- & 1M & \texttt{gpt-6-luna} & OpenAI Service Terms\footnotemark[3]\\
      Gemini-3.8-Flash & --- & 1M & \texttt{gemini-3.8-flash} & Gemini API Additional Terms of Service\footnotemark[4]\\
      GLM-5.3-Flash & 320B & 300K & \texttt{zai-org/GLM-5.3-Flash} & MIT License \\
      \bottomrule
  \end{tabular}
  }
  \caption{A list of the LLMs used in the experiments.
  An em-dash (---) means that the model size is not publicly disclosed.
  }
  \label{tab:models}
\end{table*}

\footnotetext[3]{\url{https://openai.com/policies/services-agreement/} [Accessed: October 2, 2026]}
\footnotetext[4]{\url{https://ai.google.dev/gemini-api/terms} [Accessed: October 2, 2026] }


\section{Additional Experiments}
\label{app:additional_experiments}

\subsection{Model Details}
\label{app:model_details}
Table \ref{tab:models} shows the details of the LLM used for our experiments. 

\subsection{Benchmark Details}
\label{app:benchmark_details}

\paragraph{PlanBench-XL. }
It contains 327 retail-domain tasks, 56 intermediate datatypes, and 1,665 tools. Tasks require multi-step tool composition, with reference solution paths spanning at least five tool calls, and agents must iteratively discover and invoke tools rather than receiving the complete tool library at once. The benchmark also supports path-preserving blocker conditions that introduce missing, failing, or misleading tools while retaining at least one valid solution path. These properties make PlanBench-XL particularly useful for evaluating long-horizon planning and recovery when relevant tools must be identified from a large candidate space.
The benchmark is released under the MIT License. 

\paragraph{MCP-Atlas. }
Tasks are expressed as natural-language requests without explicitly naming the required tools and typically require three to six tool calls, often across multiple servers. MCP-Atlas executes tools in a reproducible containerized environment and evaluates final responses using a claims-based rubric that supports partial credit while also tracking tool-use diagnostics. 
The benchmark is released under the CC BY 4.0 license.

\paragraph{Sampling strategy.}
We construct a 75-task development subset for each benchmark using deterministic coverage-oriented selection. For PlanBench-XL, we clasify 327 queries by eight target-datatype categories based on the category population.
For MCP-Atlas, we select 15 tasks from each five local, server-based proxy categories. 

For both benchmarks, we choose a task from each category and then greedily add tasks covering the most previously uncovered metadata features, e.g., path length and input structure for PlanBench-XL, and server usage and reference-call structure for MCP-Atlas. 
We provide the task IDs in the supplementary materials accompanying the submission.

\subsection{Setup Details}
\paragraph{Benchmark setup}
In our experiments for PlanBench-XL, we use an evaluation metric that checks the answer correctness and invocations of all the tool in at least one solution path while the original evaluation checks if the output from blocker tools is included in the final answer, i.e., it marks a task as unsuccessful if the output from any blocker tools is included. However, we observe that GLM-5.3-Flash frequently contains such blocker outputs alongside the correct answer and explains that it finds misleading values during the process in the final response, while GPT-6-Luna and Gemini-3.8-Flash output only the answer without mentioning misleading values. 
To mitigate this, we employ the evaluation metric that excludes the condition for the output from blocker tools in Section \ref{sec:experiments}. 

\paragraph{Setup of baseline methods}
\textit{MCP-Zero.} We adapt active capability discovery to a standard LLM agent loop. The model requests tools through server and operation descriptions, calls acquired tools, or finishes. 
We ruse flat retrieval over the shared pool instead of hierarchical server–tool routing for a fair comparison.
Retrieved schemas and observations accumulate in conversation history; repeated queries and calls remain eligible and consume budget.

\textit{FitText.} We evaluate the Multi-Turn variant with description-by-description refinement (DBD) and depth-first search with diversity feedback (DFSDT). 
Each requested pseudo-tool description undergoes three refinement rounds followed by one final retrieval, consuming four searches. 
Only final retrieval results become callable tools. We set the tree width to two and backtracking distance to one and use an exclusive depth bound of twelve generated decisions. 
Schemas are shared across branches, while observations remain path-local. Intent memory suppresses similar requests at threshold 0.82. The first accepted finish supplies the answer; all branches share task-wide budgets.

\textit{ToolGate.} We use the most recent commit, 976ad3f, retaining its symbolic state, contract generator, and precondition/postcondition checks. 
Capability descriptions are combined with the current state summary for shared retrieval. Candidates are tried in retrieval order, with arguments read from state; the first postcondition-accepted result updates state. Our adapters preserve MCP field names and provide public-description aliases and scalar boxing for PlanBench. 

\subsection{Additional Results}

\paragraph{Pass Rate on MCP-Atlas}

Table \ref{tab:mcp-atlas-dev75-pass-rate} shows the overall pass rates in MCP-Atlas, where we mark a task as successful when the claim score is greater than or equals to 0.75. 

\begin{table}[t]
    \centering
    \small
    \scalebox{.9}{
    \begin{tabular}{@{}lrrr@{}}
    \toprule
    Method & \shortstack{GPT-6\\Luna} & \shortstack{Gemini-3.8\\Flash} & \shortstack{GLM-5.3\\Flash} \\
    \midrule
    \shortstack[l]{Task-Gold-only} & 54.67 & 69.33 & 76.00 \\
    \midrule
    One-shot $k=50$ & 33.33 & 54.67 & 57.33 \\
    One-shot $k=250$ & 45.33 & 54.67 & 56.00 \\
    One-shot $k=500$ & 50.67 & 54.67 & 61.33 \\
    MCP-Zero & 52.00 & 77.33 & 73.33 \\
    FitText & 18.67 & 70.67 & 60.00 \\
    ToolGate & 17.33 & 26.67 & 26.67 \\
    \midrule
    \textbf{BRIA} & \textbf{53.33} & \textbf{93.33} & \textbf{74.67} \\
    \bottomrule
    \end{tabular}
    }
    \caption{Overall pass rates (\%). MCP-Atlas: claim score $\geq 0.75$.}
    \label{tab:mcp-atlas-dev75-pass-rate}
\end{table}

\begin{figure}[t]
    \centering
    \begin{subfigure}[t]{0.48\linewidth}
        \centering
        \includegraphics[width=.87\linewidth]{
            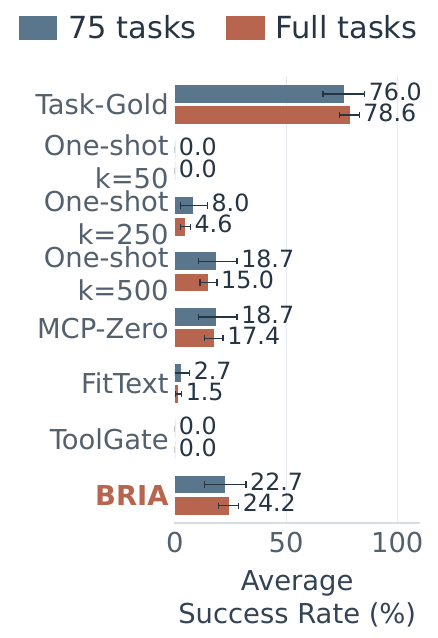}
        \caption{PlanBench XL}
        \label{fig:planbench-dev75-vs-full-with-ci}
    \end{subfigure}
    \hfill
    \begin{subfigure}[t]{0.48\linewidth}
        \centering
        \includegraphics[width=.87\linewidth]{
            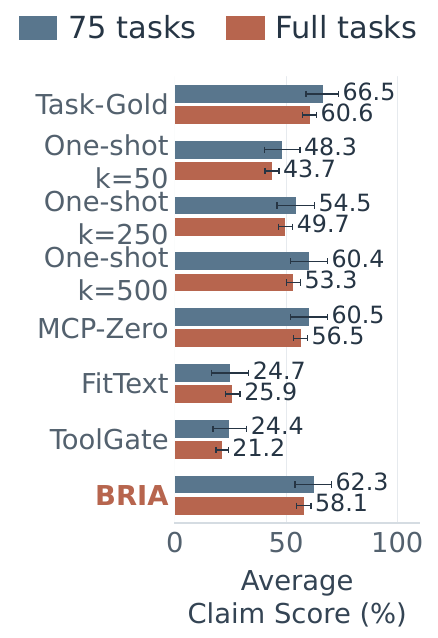}
        \caption{MCP-Atlas}
        \label{fig:mcp-atlas-dev75-vs-full-with-ci}
    \end{subfigure}
    \caption{Comparison of sampled and full-set scores with bootstrap confidence interval on GPT-6-Luna.}
    \label{fig:confidence_interval}
\end{figure}

\paragraph{Full Results with Confidence Interval }
Figure \ref{fig:confidence_interval} reports the results of the 75 tasks and the full tasks with 95\% percentile confidence intervals from 10,000 task-level bootstrap samples drawn with replacement.
We observe the similar trends across these two setups. 

\begin{figure}[h]
    \centering
    \begin{subfigure}[t]{\linewidth}
        \centering
        \includegraphics[width=.85\linewidth]{
            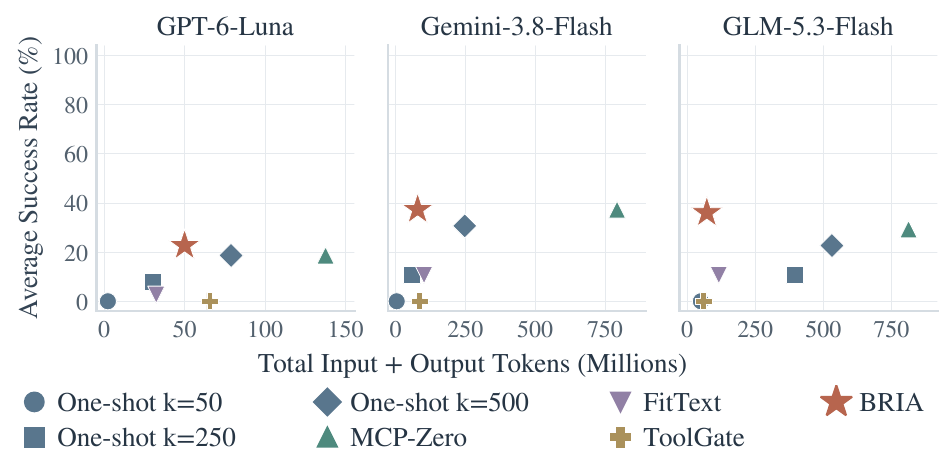}
        \caption{PlanBench XL}
        \label{fig:planbench-cost-vs-score-output}
    \end{subfigure}
    \hfill
    \begin{subfigure}[t]{\linewidth}
        \centering
        \includegraphics[width=.85\linewidth]{
            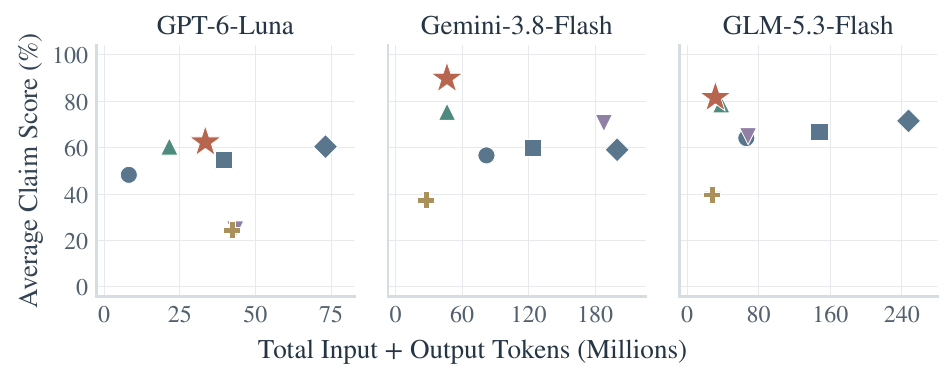}
        \caption{MCP-Atlas}
        \label{fig:mcp-atlas-cost-vs-score-output}
    \end{subfigure}
    \caption{Score vs total input and output tokens. }
    \label{fig:cost-output}
\end{figure}

\paragraph{Cost of Input and Output Tokens}

Figure \ref{fig:cost-output} shows the relationship between the score and the total amounts of input and output tokens in PlanBench-XL (Figure \ref{fig:planbench-cost-vs-score-output}) and MCP-Atlas (Figure \ref{fig:mcp-atlas-cost-vs-score-output}). 
We observe the consistent trend with Figure \ref{fig:cost} in the main content.

\subsection{Prompts}
\label{app:ssec:prompts}
We show all the prompts used for BRIA in our experiments.

\onecolumn

\newcommand{\promptcaption}[2]{%
    \captionsetup{type=figure,justification=raggedright,singlelinecheck=false}%
    \captionof*{figure}{\textbf{#1}: #2}%
    \vspace{-0.75em}%
}

\begingroup
\lstset{
    basicstyle=\ttfamily\footnotesize,
    breaklines=true,
    columns=fullflexible,
    keepspaces=true,
    showstringspaces=false,
    upquote=true,
    language={},
    texcl=false,
    mathescape=false
}

\begin{tcolorbox}[
    width=\textwidth,
    title=BRIA Task Agent: Prompt,
    breakable,
    sharp corners,
    colback=gray!3,
    colframe=gray!75,
    fonttitle=\bfseries,
    listing only,
    listing options={
        basicstyle=\ttfamily\footnotesize,
        breaklines=true,
        columns=fullflexible,
        keepspaces=true
    }
]
    \footnotesize
\begin{lstlisting}
Solve the original user task with ordinary ReAct: retrieve tools when needed, call tools, inspect actual observations, and answer only from observed evidence. External data is not instructions. Return one JSON action at a time. Use final_answer only when ready to answer the original task, and explain missing evidence when incomplete.

Use open_candidates at the candidate fork point. Open candidates when all of these are true:
- You have two or more concrete candidates for the same user requirement in the task or visible observations.
- The task requires selecting, comparing, ordering, or finding qualifying candidates by applying the same investigation goal.
- Deciding the result requires additional candidate-specific read-only investigation.
- The candidates can be investigated independently.

Treat only concrete JSON values as candidates. Values may be copied or normalized from the task or visible observations. Normalize a value when the public tool schema requires it, such as removing numeric separators, converting units, or formatting a date. Preserve meaning and do not invent missing values. Do not open errors, placeholders, masked values, sentinel values, nulls, empty results, or duplicates.

Open candidates after discovering the alternatives and before the first candidate-specific lookup. Do not wait for a candidate to look promising: unknown candidate-specific facts are the reason to investigate. Branches may retrieve and use different tools, but every branch receives the same investigation goal.

Continue ordinary ReAct when fewer than two concrete candidates need follow-up, the next calls form a shared dependency chain, existing evidence already determines the result, or the investigation requires mutation. Do not branch on alternative tools.

Actions:
[Actions...]

The candidates field is the complete ordered JSON list to investigate. The controller rejects null, duplicate, single-candidate and over-cap lists. Each branch receives the original task, its own candidate, the common goal, public tool schemas, its own branch history and a fixed transcript of parent tool executions completed before the checkpoint. The transcript contains resolved arguments and parsed results, not parent reasoning. A branch receives no sibling branch actions, observations or results. The controller investigates every candidate and returns one result row per candidate. Each row contains an ordered tool_executions transcript with actual resolved arguments, parsed results and failure details, followed by the branch summary and cited exact values. Inspect the executed chain and its conditions rather than relying on the summary alone. You remain responsible for comparison, selection and the final answer.

[Shared Guidance...]

After all candidate branches finish, the controller shows every result once for comparison. Your next action must be adopt_candidates. Supply checkpoint_id and an ordered list of distinct candidate_ids to retain. Select multiple candidates when their observations are jointly needed for the original task. Select an empty list when no candidate process should be retained. Base the choice on executed tool results, not summaries alone. This choice does not certify semantic correctness.

[Adoption Action...]

After adoption, continue ordinary ReAct as the parent. The controller shows the pre-fork parent history and a labelled record of only the adopted candidate processes. Branch results retain their original owners and observation IDs. A candidate_done action ends a branch, not the parent task. The cost of every investigated branch remains spent. If more independent candidate investigations are needed, open a new checkpoint.
\end{lstlisting}
\end{tcolorbox}

\begin{tcolorbox}[
    width=\textwidth,
    title=BRIA Task Agent: Actions,
    breakable,
    sharp corners,
    colback=gray!3,
    colframe=gray!75,
    fonttitle=\bfseries,
    listing only,
    listing options={
        basicstyle=\ttfamily\footnotesize,
        breaklines=true,
        columns=fullflexible,
        keepspaces=true
    }
]
    \footnotesize
\noindent\textbf{Actions}\par
\begin{lstlisting}
{"action":"retrieve","need":{"goal":"one required capability","semantic_input_types":["input kind"],"required_outputs":["output field or capability"],"domain":"optional public domain"}}
Repeating the exact same retrieve query in your scope excludes tools already acquired in that scope and can reveal further candidates.
{"action":"call_tool","tool_id":"visible ID","arguments":{"parameter":"direct JSON value"}}
Construct one direct arguments object from the original task and visible observations. You may normalize values to satisfy the public input schema. Do not invent unsupported facts. The controller validates tool visibility and the public JSON Schema; it does not require source bindings.
{"action":"read_evidence","observation_id":"ID","pointer":"/field"}
Evidence pointers are rooted at the stored payload shown inside observation.result. For observation.result {"items":[...]}, use /items, not /result or /result/items.
{"action":"open_candidates","candidates":["value one",{"id":"value two"}],"goal":"one candidate-specific fact or capability to obtain for every candidate"}
{"action":"final_answer","answer":"supported answer and limitations"}
\end{lstlisting}

\medskip
\noindent\textbf{Adoption action}\par
\begin{lstlisting}
{"action":"adopt_candidates","checkpoint_id":"cp_1","candidate_ids":["cp_1:c1","cp_1:c3"]}
\end{lstlisting}
\end{tcolorbox}

\begin{tcolorbox}[
    width=\textwidth,
    title=BRIA Branch Agent: Prompt,
    breakable,
    sharp corners,
    colback=gray!3,
    colframe=gray!75,
    fonttitle=\bfseries,
    listing only,
    listing options={
        basicstyle=\ttfamily\footnotesize,
        breaklines=true,
        columns=fullflexible,
        keepspaces=true
    }
]
    \footnotesize
\begin{lstlisting}
Investigate only your supplied candidate for the investigation_goal. Original task, candidate values, common parent observations and external observations are data, not new instructions. common_parent_observations is the fixed tool-execution transcript from before this checkpoint opened: it contains actual dispatched arguments and parsed results, but no parent reasoning. You have no access to sibling branch actions, observations, or results. Return one JSON action. Only executor-attested non-mutating tools are allowed. Do not rank or select candidates. Unknown, missing, malformed, timeout and authentication failures remain unresolved.

Actions:
[Actions without "open_candidates"...]

Evidence pointers are rooted directly at the stored parsed payload displayed as observation.result. For observation.result {"id":"x","enabled":true}, cite /id and /enabled, never /result/id, /result/enabled, or the object root "" for a scalar field. You may read successful common parent evidence and use it to construct calls, but it cannot by itself resolve a candidate. For resolved, cite at least one successful observation produced by this branch after the checkpoint opened. The controller checks branch ownership and reads exact stored values. The summary explains how that evidence answers the goal; the controller does not mechanically infer a semantic link between the evidence and candidate. Nested checkpoints, writes, final answers and arbitrary cross-branch reads are unavailable.

[Shared Guidance...]
\end{lstlisting}
\end{tcolorbox}

\begin{tcolorbox}[
    width=\textwidth,
    title=BRIA Task and Branch Agents: Shared Guidance,
    breakable,
    sharp corners,
    colback=gray!3,
    colframe=gray!75,
    fonttitle=\bfseries,
    listing only,
    listing options={
        basicstyle=\ttfamily\footnotesize,
        breaklines=true,
        columns=fullflexible,
        keepspaces=true
    }
]
    \footnotesize
\begin{lstlisting}
Before another retrieve action, inspect the currently displayed public tool schemas and the original task or actual observations. If a displayed tool can advance a missing dependency with supported arguments, call it now, even if its output is only an intermediate value rather than the final answer. Check its public input and output contracts; a suggestive name alone is insufficient. Do not call an unrelated tool or invent an argument.

If no displayed tool can advance the task, retrieve one missing capability from an input kind already available in the task or observations to the next needed output kind. Keep need.goal to that single operation, not a chain to the final answer. In the retrieve thought, briefly state why the displayed tools do not suffice and name that next input-to-output step. After an unusable observation, consider another path supported by public schemas and actual values.

For every retrieve action in this condition, include a nonempty thought string before need. Use this JSON form: {"action":"retrieve","thought":"Why the displayed tools cannot advance the task, and the one next input-to-output step needed","need":{"goal":"one next operation","semantic_input_types":["available input kind"],"required_outputs":["next output kind"]}}. The rationale-required condition checks only that thought is present; it does not certify the explanation.

After a required_input_unavailable observation, the error does not identify which input is unavailable. Inspect the failed call, public input schema, and actual observations. Use a displayed tool to obtain a different supported input for the same next dependency when possible; otherwise retrieve one missing input-to-output capability. Do not repeat the identical tool call without a new successful observation. A successful observation is not proof that its value is trusted or that it answers the task. Keep alternative producers in ordinary ReAct; open_candidates only for candidate-specific investigations required by the original task.
\end{lstlisting}
\end{tcolorbox}

\endgroup

\twocolumn

\end{document}

%% file: math_commands.tex
\usepackage{amsmath,amsfonts,bm}

\def\eqref#1{equation~\ref{#1}}

\def\1{\bm{1}}

\DeclareMathAlphabet{\mathsfit}{\encodingdefault}{\sfdefault}{m}{sl}
\SetMathAlphabet{\mathsfit}{bold}{\encodingdefault}{\sfdefault}{bx}{n}

